\documentclass{article}
\usepackage{ijcai20}
\usepackage{times}
\usepackage{soul}
\usepackage{url}
\usepackage[hidelinks]{hyperref}
\usepackage[utf8]{inputenc}
\usepackage[small]{caption}
\usepackage{graphicx}
\usepackage{amsmath}
\usepackage{amsthm}
\usepackage{booktabs}
\usepackage{smartdiagram}
\usepackage{tikz}

\usetikzlibrary{arrows.meta}
\tikzset{%
	>={Latex[width=2mm,length=2mm]},
	base/.style = {rectangle, rounded corners, draw=black,
		minimum width=0.2cm, minimum height=1cm,
		text centered, font=\sffamily},
	activityStarts/.style = {base},
	nlp/.style = {base, fill=red!30},
	ml/.style = {base, fill=green!30},
	diagonal fill/.style 2 args={fill=#2, path picture={
			\fill[#1, sharp corners] (path picture bounding box.south west) -|
			(path picture bounding box.north east) -- cycle;}},
	reversed diagonal fill/.style 2 args={fill=#2, path picture={
			\fill[#1, sharp corners] (path picture bounding box.north west) |- 
			(path picture bounding box.south east) -- cycle;}},
	dl/.style = {base, fill=blue!30}
}
\title{Relation Clustering in Narrative Knowledge Graphs}
\author{Simone Mellace \and Vani K \And Alessandro Antonucci\footnote{Contact Author}
\affiliations IDSIA - Lugano (Switzerland)\\
\emails\{simone,vanik,alessandro\}@idsia.ch}
\begin{document}
\maketitle
\begin{abstract}
When coping with literary texts such as novels or short stories, the extraction of structured information in the form of a knowledge graph might be hindered by the huge number of possible relations between the entities corresponding to the characters in the novel and the consequent hurdles in gathering supervised information about them. Such issue is addressed here as an unsupervised task empowered by transformers: relational sentences in the original text are embedded (with SBERT) and clustered in order to merge together semantically similar relations. All the sentences in the same cluster are finally summarized (with BART) and a descriptive label extracted from the summary. Preliminary tests show that such clustering might successfully detect similar relations, and provide a valuable preprocessing for semi-supervised approaches.
\end{abstract}

\section{Introduction}
Recent applications in the field of \emph{Natural Language Processing} (NLP) are exploiting data-driven techniques from the general area of \emph{Machine Learning} (ML). These are typically \emph{Deep Learning} (DL) systems based on multi-layer neural networks fitted with the input text data, to be converted in numerical objects by some embedding scheme. Such DL-NLP systems are successful in extracting knowledge from natural language and capturing the underlying \emph{narratives}.

As a matter of fact, most of these NLP efforts are focused on a few mainstream applicative areas, such as biomedical literature \cite{zhang2018hybrid,lv2016clinical} or news and social media  \cite{trieu2017news,ghosh2018towards}. Other inputs such as \emph{literary text} in the form of novels or short stories received less attention \cite{wohlgenannt2016extracting,volpetti2020a}. This is unfortunate as literary texts might exhibit high complexity in the narrative plots, while also lacking explicit annotations, thus making the knowledge extraction process very challenging. Handling such complexities, helps in evaluating the models Natural Language Understanding and creating benchmarks for these low-resource domains. This could in turn be helpful for common sense reasoning, reading comprehensions and enhance NLP applications such as summary generation, machine translations and question answering.

Despite their astonishing applications in NLP, e.g.,  \cite{zhu2019joint,paulus2017deep}, DL models are typically based on discriminative functions with a huge number of parameters, whose interpretation is often problematic. This prevents both the \emph{explainability} of the results and the possibility of doing \emph{reasoning} over the model entities. For this reason, alternative approaches to NLP, based on so-called \emph{Knowledge Graphs} (KGs), i.e., relational ontologies providing interlinked descriptions of the entities involved in a text, are also popular. Despite the existence of techniques for automatic KG extraction acting at the syntactic level \cite{tang2016recurrent,ruan2016automatic}, most of the approaches require  supervision in the form of manual annotations or access to knowledge bases, such as UMLS\footnote{\url{https://uts.nlm.nih.gov}}, for higher level descriptions.

Of course these two orthogonal perspectives, say DL and KGs, can be combined. DL models can be trained from KGs \cite{socher2013reasoning,li2019knowledge} and used for ML, and, \emph{vice versa}, DL models such as embeddings can be used to predict missing links of the KG, classify relations, or align entities from different KGs \cite{liu2019k,lin2015learning}.

Here, we follow such an integrated point of view, being motivated by specific features of literary text understanding. In fact, for this kind of text, the KG entities are typically the characters in the plot, and no serious alignment issues appear, while the classification of the relations becomes much more challenging because of the lack of supervision. In other domains the number of possible relations is typically limited (e.g., in \cite{chen2010chem2bio2rdf}, few relations such as \emph{binding}, \emph{expression}, \emph{protein interaction} and few others), while in the literary case the possible relations between characters (e.g., Table \ref{tab:sentences}) can be much more.
Accordingly, we explore some directions for an unsupervised approach to the identification of relations in KGs obtained from literary texts. The goal is to cluster semantically equivalent relational sentences including descriptions of relations between the characters of a novel. Our preliminary tests seem to be promising with respect to the proper identification of similar relations, while also giving directions about the most suitable clustering strategies as well as further development of semi-supervised tools.

The paper is organized as follows. In Section \ref{sec:work}, we summarize the existing literature in the field. Section \ref{sec:flow} describes our  workflow, which is demonstrated by applicative examples in Section \ref{sec:exp}. Conclusions and outlooks are in Section \ref{sec:conc}.

\section{Existing Work}\label{sec:work}
As discussed in the previous section, DL tools such as sequence and self-attention models as well as transformers (e.g., BERT, Xlnet, BART) have been widely and successfully used in NLP for word and sentence encoding \cite{peters2018deep,devlin2018bert,yang2019xlnet,lewis2019bart}. These models can be fine-tuned and used for various tasks such as classification, summarization and sentiment analysis. This also concerns KGs, where DL models are used for embedding the triplet information and used for tasks such as link predictions and KG completion \cite{lin2019kagnet,yao2019kg}, while other researchers worked on the training of embedding from KGs \cite{ji2015knowledge,wang2014knowledge,lin2015learning,bordes2013translating}.

None of these application was concerned with literary text. Despite some attempts to apply ML and DL models in the field
\cite{worsham2018genre,short2019text,labatut2019extraction,vani2019novel2graph,volpetti2020a} to analyze character relations, sentiments and visualizations, a connection with KGs still remains under-explored. The goal of this paper is to fill this gap by providing an unsupervised alternative to the \emph{relational} classifiers recently developed for supervised tasks in \cite{vani2020temporal}.

\section{Workflow}\label{sec:flow}
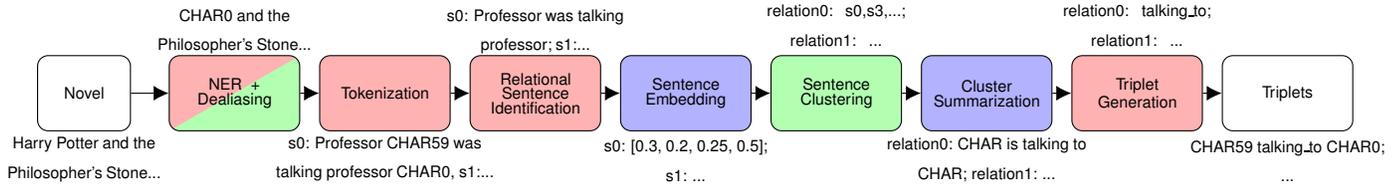
\begin{figure*}[htp!]
	\begin{tikzpicture}[node distance=1cm,
	every node/.style={fill=white, font=\sffamily}, align=center]
	\node (start)[activityStarts, xshift=-2cm, text width=1cm, label={[yshift=-1.8cm, text width=2.1cm, align=flush center]\tiny Harry Potter and the Philosopher's Stone...}]{\tiny  Novel \baselineskip=2pt \par};
	\node (onCreateBlock)[diagonal fill={green!30}{red!30},
	text width=1.5cm,
	rectangle, rounded corners, draw=black,
	minimum width=1cm, minimum height=1cm,
	text centered, font=\sffamily, right of=start, xshift=1cm, label={[text width=2.1cm, align=flush center, yshift=-0.1cm]\tiny CHAR0 and the Philosopher's Stone...}]{\tiny NER + Dealiasing \baselineskip=2pt \par};
	\node (onStartBlock)[nlp, text width=1.5cm, right of=onCreateBlock, xshift=1cm, label={[yshift=-1.8cm, text width=3cm, align=flush center]\tiny s0{:} Professor CHAR59 was talking professor CHAR0, s1{:}...}]{\tiny Tokenization \baselineskip=2pt \par};
	\node (onResumeBlock)[nlp, text width=1.5cm, right of=onStartBlock, xshift=1cm, label={[text width=3cm, align=flush center, yshift=-0.1cm]\tiny s0{:} Professor was talking professor; s1{:}...}]{\tiny Relational Sentence Identification \baselineskip=4pt \par};
	\node (onResumeBlock2)[dl, text width=1.5cm, right of=onResumeBlock, xshift=1cm, label={[text width=2.5cm, yshift=-1.8cm, align=flush center]\tiny s0{:} [0.3, 0.2, 0.25, 0.5];  s1{:} ...}] {\tiny Sentence Embedding \baselineskip=2pt \par};
	\node (activityRuns)[ml, text width=1.5cm, right of=onResumeBlock2, xshift=1cm, label={[text width=2cm]\tiny relation0{:} s0,s3,...; relation1{:} ...}] {\tiny Sentence Clustering \baselineskip=2pt \par};
	\node (onPauseBlock)[dl, text width=1.5cm, right of=activityRuns, xshift=1cm, label={[text width=3cm, yshift=-1.8cm, align=flush center]\tiny relation0{:} CHAR is talking to CHAR; relation1{:} ...}]  {\tiny Cluster Summarization \baselineskip=2pt \par};
	\node (onStopBlock)       [nlp, text width=1.5cm, right of=onPauseBlock, xshift=1cm, label={[text width=2cm]\tiny relation0{:} talking\_to; relation1{:} ...}]    {\tiny Triplet Generation \baselineskip=2pt \par};
	\node (onDestroyBlock)[activityStarts, text width=1.5cm, right of=onStopBlock, xshift=1cm, label={[yshift=-1.8cm, text width=2.6cm, align=flush center]\tiny CHAR59 talking\_to CHAR0; ...}]{\tiny Triplets \baselineskip=2pt \par};
	\draw[->]             (start) -- (onCreateBlock);
	\draw[->]     (onCreateBlock) -- (onStartBlock);
	\draw[->]      (onStartBlock) -- (onResumeBlock);
	\draw[->]     (onResumeBlock) -- (onResumeBlock2);
	\draw[->]     (onResumeBlock2) -- (activityRuns);
	\draw[->]      (activityRuns) -- (onPauseBlock);
	\draw[->]      (onPauseBlock) -- (onStopBlock);
	\draw[->]       (onStopBlock) -- (onDestroyBlock);
	\end{tikzpicture}
	\caption{The workflow of the triplet generation process, with relation clustering.\label{fig:flow}}
\end{figure*}

Figure \ref{fig:flow} depicts the workflow of the approach we propose for the unsupervised identification of similar relations in the KGs obtained from literary text. This involves a NLP part for preprocessing (entity recognition, sentence tokenization, detection of relational sentences and triplet generation) corresponding to the red blocks, a DL abstraction level (sentence embedding and summarization) corresponding to the blue blocks, as well as classical ML techniques (characters de-aliasing, sentence clustering, semi-supervised extension) associated with green blocks. These steps in their sequential order are described here below together with the main challenges they present. The tool is available as a free software.\footnote{https://github.com/IDSIA/novel2graph}

\paragraph{Named Entity Recognition (NER).} The very first step is the identification of the entities to be associated with the KG nodes. These are detected by a custom version of the Stanford NER Tagger\footnote{\url{https://nlp.stanford.edu/software/CRF-NER.html}} such that consecutive entities in a sentence (i.e. words tagged as \textit{PERSON}), are detected as a unique element (e.g., \emph{Harry James Potter}). 

\paragraph{Dealiasing.} As a same character can be termed with different \emph{aliases} in the same novel, a de-aliasing might be required. This issue has been already addressed in \cite{vani2020temporal}, where a satisfactory solution based on ML and NLP has been found. Here we adopt a similar strategy based on the classical DBSCAN clustering  ($\epsilon=.3$ and Levenshtein string distances), together with a number of manual adjustments. In our approach in fact, we first perform separate pre-clustering over entities  starting with the same letter (e.g., \emph{Hermione} and \emph{Hermione Granger} are identified as a cluster while \emph{Harry}, \emph{Harry Potter} and \emph{H. Potter} as another one), and then adding similar but unassigned names to a cluster (e.g., \emph{Granger} assigned to \emph{Hermione}'s cluster and \emph{Potter} to \emph{Harry}). All the occurrences of the aliases in the same cluster are finally replaced by identifiers (e.g., CHAR0 replaces \emph{Harry}, \emph{Potter}, \emph{Harry Potter} and so on).

\paragraph{Tokenization.} Embeddings based on tranformers are based on contextual information. Since, sentences are considered as the simplest logical and meaningful unit that provides a semantic intuition of the context, we rely on a segmentation at this level.

\paragraph{Relational Sentence Identification.} Let us call \emph{relational} a sentence including two or more characters. We extract relational sentences from the de-aliased and tokenized text, by also evaluating whether or not the text between the two character occurrences is a simple proposition or not (e.g., \emph{Harry and Ron were having good time} and \emph{Harry looked at Ron}). 
If this is the case we call the relation \emph{symmetric} and we generate two distinct input for the pipeline. Note also we only use sentences containing exactly two characters and excluding self-relations (e.g., \emph{Harry, I am Harry Potter}).

\paragraph{Sentence Embedding.} To identify the relations between entities, we embed the relational sentences using Sentence BERT (SBERT) \cite{reimers2019sentence}. SBERT uses a Siamese network structure \cite{schroff2015facenet} to reproduce meaningful encodings. The method was specifically modelled for clustering and semantic search. SBERT adds a pooling operation on top of BERT to derive these embeddings. SBERT is fine tuned on SNLI \cite{bowman2015large} and MNLI \cite{williams2018broad} datasets with a three-way soft-max classifier objective function for one epoch with the default pooling strategy MEAN (computing the mean of all output vectors).

\paragraph{Sentence Clustering.} 
Since, these embeddings encode semantic and contextual information, sentences with similar vector representations are supposed to share similar relations. Hence, we adopt a simple clustering approach to group the sentences with similar relations. The distances between the vectors returned by SBERT are assumed to reflect the semantic similarity between the corresponding sentences and hence the relations included in these sentences. Classical clustering methods such as k-means or DBSCAN can be therefore used to create groups of sentences and hence triplets with the same relation. We considered the Euclidean distance, as well as the classical cosine distance. Even though clustering the entire sentence may not explicitly cluster the relationships, the sentences that fall into similar semantic spaces can provide us a coarse-grained grouping of  relations. 

\paragraph{Cluster Summarization.} After the clustering of the relational sentences, we might want to represent these relations as a summary of the sentences involved in the cluster. To achieve that we adopt the BERT summarization pipeline based on the BART \cite{lewis2019bart} model. This includes an encoder like BERT and a decoder like GPT \cite{radford2019language} and it is trained on CNN/Daily Mail dataset with learning rate $3\cdot 10^{-5}$ (Adam optimizer). This performs extractive summarization, giving most suitable representative sentences of each cluster. Although training is not in-domain, as news articles are also narratives, we use this for a preliminary set-up.

\paragraph{Triplet Generation.} Once the extractive summary is produced, for asymmetric relations we extract the phrase which comes between the two reference characters. We then extract only the verbs from these phrases, which are the considered as part-of-speech tags that could convey some information about the type of relations. 

\paragraph{From Unsupervised to Semi-Supervised Learning.} The overall procedure described in this section is purely unsupervised. Yet, the clusters of relational sentences are described by the summaries, first, and then labels generated by the system. This might be the basis of a system where, part of those clusters are manually inspected and their summaries/labels validated or fixed by a human annotator. This would turn the system into a semi-supervised one, where the annotated clusters can be used as classifier of the relations.

\begin{table}[htp!]
    {\centering
        \tiny
        \begin{tabular}{lp{7cm}}
            \hline
            Book&Sentence\\
            \hline
            \hline
            HP&\emph{Dumbledore} smiled at the look of amazement on \emph{Henry}'s face\\
            HP&\emph{Ron} grinned at \emph{Henry}\\
            LW&\emph{Brooke} smiling at \emph{Meg} as if everything had become possible him now\\
            HP&\emph{Henry} stared as \emph{Dumbledore} sidled back into the picture $\ldots$ gave him a small smile\\
            \hline
    \end{tabular}}
    \caption{Relational sentences from the same cluster.}
    \label{tab:sentences}
\end{table}
 
\section{Experiments}\label{sec:exp}
\begin{figure}[htp!]
\centering
\definecolor{saidcolor}{rgb}{0.27,0.23,0.21}
\definecolor{smilecolor}{rgb}{0.18,0.68,0.02}
\definecolor{lookcolor}{rgb}{0.04,0.61,0.8}
\definecolor{seecolor}{rgb}{0.05,0.4,0.96}
\definecolor{askcolor}{rgb}{0.22,0.42,0.13}
\definecolor{walkcolor}{rgb}{0.71,0.21,0.5}
\begin{tikzpicture}[scale=0.12,
  blacknode/.style={shape=circle, draw=black, line width=2},
  bluenode/.style={shape=circle, draw=blue, line width=2},
  greennode/.style={shape=circle, draw=green, line width=2},
  rednode/.style={shape=circle, draw=red, line width=2},
  saidnode/.style={shape=rectangle,fill=saidcolor,  line width=2},
  smilenode/.style={shape=rectangle,fill=smilecolor, line width=2},
  looknode/.style={shape=rectangle, fill=lookcolor, line width=2},
  seenode/.style={shape=rectangle, fill=seecolor, line width=2},
  asknode/.style={shape=rectangle, fill=askcolor, line width=2},
  walknode/.style={shape=rectangle, fill=walkcolor, line width=2},
  emptynode/.style={shape=rectangle, draw=black, line width=2}
]
\node (Albus Dumbledore) at (156.17bp,1467.5bp) [minimum size=0.1cm, draw,circle] {\tiny Dumbledore};
\node (Amy) at (1891.2bp,1452.0bp) [minimum size=1.5cm,draw,circle] {\tiny Amy};
\node (Jo) at (1533.2bp,1067.5bp) [minimum size=1.5cm, draw,circle] {\tiny Jo};
\node (Laurence) at (1108.2bp,601.0bp) [minimum size=1.8cm, draw,circle] {\tiny Laurence};
\node (Beth) at (1033.2bp,165.5bp) [draw,circle] {\tiny Beth};
\node (H Potter) at (685.17bp,1452.0bp) [minimum size=1.8cm, draw,circle] {\tiny Harry};
\node (Hagar) at (135.17bp,767.5bp) [draw,circle] {\tiny Hagar};
\node (Hermione) at (324.17bp,155.5bp) [draw,circle] {\tiny Hermione};
\node (Ron) at (585.17bp,701.0bp) [minimum size=1.5cm, draw,circle] {\tiny Ron};
\node (Snape) at (1047.2bp,1067.5bp) [draw,circle] {\tiny Snape};
\node (Meg) at (1730.2bp,601.0bp) [minimum size=1.5cm, draw,circle] {\tiny Meg};
\definecolor{strokecolor}{rgb}{0.27,0.23,0.21};
\draw [strokecolor,->] (Amy) ..controls (1635.5bp,1200.3bp) and (1646.1bp,1194.9bp)  .. (Jo);
\definecolor{strokecolor}{rgb}{0.05,0.4,0.96};
\draw [strokecolor,->] (Amy) ..controls (1672.9bp,1200.3bp) and (1663.7bp,1195.9bp)  .. (Jo);
\definecolor{strokecolor}{rgb}{0.62,0.42,0.93};
\draw [strokecolor,->] (Amy) ..controls (1299.4bp,1343.0bp) and (1290.8bp,1302.1bp)  .. (1283.2bp,1264.0bp) .. controls (1255.6bp,1125.8bp) and (1227.4bp,970.82bp)  .. (Laurence);
\definecolor{strokecolor}{rgb}{0.24,0.01,0.43};
\draw [strokecolor,->] (Amy) ..controls (1316.4bp,1338.2bp) and (1308.3bp,1299.9bp)  .. (1301.2bp,1264.0bp) .. controls (1273.5bp,1125.5bp) and (1245.2bp,970.19bp)  .. (Laurence);
\definecolor{strokecolor}{rgb}{0.65,0.93,0.38};
\draw [strokecolor,->] (Amy) ..controls (1351.1bp,1331.8bp) and (1343.7bp,1296.9bp)  .. (1337.2bp,1264.0bp) .. controls (1309.2bp,1124.0bp) and (1280.6bp,966.74bp)  .. (Laurence);
\definecolor{strokecolor}{rgb}{0.18,0.69,0.49};
\draw [strokecolor,->] (Amy) ..controls (1368.8bp,1330.4bp) and (1361.6bp,1296.2bp)  .. (1355.2bp,1264.0bp) .. controls (1326.9bp,1122.4bp) and (1298.0bp,963.29bp)  .. (Laurence);
\definecolor{strokecolor}{rgb}{0.24,0.56,0.64};
\draw [strokecolor,->] (Beth) ..controls (1824.5bp,257.9bp) and (1995.5bp,157.34bp)  .. (Amy);
\definecolor{strokecolor}{rgb}{0.31,0.91,0.92};
\draw [strokecolor,->] (Beth) ..controls (1791.0bp,233.97bp) and (2100.2bp,147.37bp)  .. (Amy);
\definecolor{strokecolor}{rgb}{0.33,0.35,0.17};
\draw [strokecolor,->] (Beth) ..controls (1497.8bp,479.01bp) and (1497.1bp,701.32bp)  .. (Jo);
\definecolor{strokecolor}{rgb}{0.31,0.91,0.92};
\draw [strokecolor,->] (Beth) ..controls (1512.0bp,479.9bp) and (1511.6bp,700.55bp)  .. (Jo);
\definecolor{strokecolor}{rgb}{0.92,0.95,0.18};
\draw [strokecolor,->] (Beth) ..controls (1540.2bp,480.25bp) and (1540.4bp,700.25bp)  .. (Jo);
\definecolor{strokecolor}{rgb}{0.42,0.21,0.05};
\draw [strokecolor,->] (Beth) ..controls (1554.4bp,479.9bp) and (1554.8bp,700.55bp)  .. (Jo);
\definecolor{strokecolor}{rgb}{0.16,0.67,0.31};
\draw [strokecolor,->] (Beth) ..controls (1568.5bp,479.01bp) and (1569.2bp,701.32bp)  .. (Jo);
\definecolor{strokecolor}{rgb}{0.62,0.73,0.45};
\draw [strokecolor,->] (Beth) ..controls (1582.7bp,477.39bp) and (1583.7bp,702.25bp)  .. (Jo);
\definecolor{strokecolor}{rgb}{0.04,0.61,0.8};
\draw [strokecolor,->] (H Potter) ..controls (427.61bp,1360.1bp) .. (Albus Dumbledore);
\definecolor{strokecolor}{rgb}{0.13,0.64,0.69};
\draw [strokecolor,->] (H Potter) ..controls (308.39bp,1289.6bp) .. (Hagar);
\definecolor{strokecolor}{rgb}{0.22,0.42,0.13};
\draw [strokecolor,->] (H Potter) ..controls (330.18bp,1278.2bp).. (Hagar);
\definecolor{strokecolor}{rgb}{0.71,0.21,0.5};
\draw [strokecolor,->] (H Potter) ..controls (373.27bp,1256.4bp) .. (Hagar);
\definecolor{strokecolor}{rgb}{0.41,0.3,0.3};
\draw [strokecolor,->] (H Potter) ..controls (786.43bp,1300.1bp) and (792.73bp,1281.9bp)  .. (798.17bp,1264.0bp) .. controls (899.94bp,929.63bp) and (911.52bp,514.12bp)  .. (Hermione);
\definecolor{strokecolor}{rgb}{0.63,0.97,0.38};
\draw [strokecolor,->] (H Potter) ..controls (802.84bp,1304.6bp) and (810.06bp,1284.1bp)  .. (816.17bp,1264.0bp) .. controls (917.94bp,929.63bp) and (929.52bp,514.12bp)  .. (Hermione);
\definecolor{strokecolor}{rgb}{0.26,0.84,0.16};
\draw [strokecolor,->] (H Potter) ..controls (592.31bp,1150.7bp) and (594.16bp,940.88bp)  .. (Ron);
\definecolor{strokecolor}{rgb}{0.41,0.3,0.3};
\draw [strokecolor,->] (H Potter) ..controls (621.08bp,1147.5bp) and (622.3bp,945.13bp)  .. (Ron);
\definecolor{strokecolor}{rgb}{0.18,0.37,0.97};
\draw [strokecolor,->] (H Potter) ..controls (649.64bp,1145.8bp) and (650.29bp,948.24bp)  .. (Ron);
\definecolor{strokecolor}{rgb}{0.22,0.42,0.13};
\draw [strokecolor,->] (H Potter) ..controls (706.48bp,1145.4bp) and (706.1bp,948.96bp)  .. (Ron);
\definecolor{strokecolor}{rgb}{0.56,0.49,0.71};
\draw [strokecolor,->] (H Potter) ..controls (720.71bp,1145.8bp) and (720.06bp,948.24bp)  .. (Ron);
\definecolor{strokecolor}{rgb}{0.79,0.72,0.99};
\draw [strokecolor,->] (H Potter) ..controls (734.97bp,1146.6bp) and (734.04bp,946.74bp)  .. (Ron);
\definecolor{strokecolor}{rgb}{0.56,0.25,0.5};
\draw [strokecolor,->] (H Potter) ..controls (778.03bp,1150.7bp) and (776.18bp,940.88bp)  .. (Ron);
\definecolor{strokecolor}{rgb}{0.15,0.4,0.69};
\draw [strokecolor,->] (H Potter) ..controls (855.31bp,1320.0bp) and (877.02bp,1291.5bp)  .. (897.17bp,1264.0bp) .. controls (919.93bp,1233.0bp) and (943.7bp,1198.3bp)  .. (Snape);
\definecolor{strokecolor}{rgb}{0.13,0.64,0.69};
\draw [strokecolor,->] (H Potter) ..controls (868.8bp,1325.9bp) and (892.94bp,1294.3bp)  .. (915.17bp,1264.0bp) .. controls (935.96bp,1235.6bp) and (957.6bp,1204.3bp)  .. (Snape);
\definecolor{strokecolor}{rgb}{0.41,0.55,0.47};
\draw [strokecolor,->] (Hagar) ..controls (345.78bp,1006.8bp).. (H Potter);
\definecolor{strokecolor}{rgb}{0.07,0.04,0.05};
\draw [strokecolor,->] (Hagar) ..controls (386.79bp,1016.3bp).. (H Potter);
\definecolor{strokecolor}{rgb}{0.38,0.09,0.35};
\draw [strokecolor,->] (Hagar) ..controls (426.53bp,1026.7bp).. (H Potter);
\definecolor{strokecolor}{rgb}{0.43,0.41,0.69};
\draw [strokecolor,->] (Hagar) ..controls (466.04bp,1036.2bp).. (H Potter);
\definecolor{strokecolor}{rgb}{0.0,0.11,0.32};
\draw [strokecolor,->] (Hagar) ..controls (394.64bp,819.21bp)  .. (Ron);
\definecolor{strokecolor}{rgb}{0.82,0.29,0.23};
\draw [strokecolor,->] (Hermione) ..controls (947.9bp,499.35bp) and (937.75bp,923.7bp)  .. (834.17bp,1264.0bp) .. controls (828.2bp,1283.6bp) and (821.18bp,1303.6bp)  .. (H Potter);
\definecolor{strokecolor}{rgb}{0.47,0.64,0.75};
\draw [strokecolor,->] (Hermione) ..controls (295.7bp,311.24bp) .. (Hagar);
\definecolor{strokecolor}{rgb}{0.62,0.42,0.93};
\draw [strokecolor,->] (Jo) ..controls (1691.5bp,1148.8bp) and (1801.0bp,1224.7bp)  .. (Amy);
\definecolor{strokecolor}{rgb}{0.24,0.67,0.93};
\draw [strokecolor,->] (Jo) ..controls (1694.1bp,1132.3bp) and (1835.6bp,1228.2bp)  .. (Amy);
\definecolor{strokecolor}{rgb}{0.33,0.35,0.17};
\draw [strokecolor,->] (Jo) ..controls (1482.9bp,715.2bp) and (1483.4bp,490.44bp)  .. (Beth);
\definecolor{strokecolor}{rgb}{0.09,0.68,0.04};
\draw [strokecolor,->] (Jo) ..controls (1526.0bp,713.12bp) and (1526.1bp,493.5bp)  .. (Beth);
\definecolor{strokecolor}{rgb}{0.38,0.23,0.11};
\draw [strokecolor,->] (Jo) ..controls (1362.7bp,898.6bp) and (1334.9bp,860.84bp)  .. (Laurence);
\definecolor{strokecolor}{rgb}{0.82,0.4,0.09};
\draw [strokecolor,->] (Jo) ..controls (1406.5bp,866.27bp) and (1381.4bp,830.09bp)  .. (Laurence);
\definecolor{strokecolor}{rgb}{0.18,0.69,0.49};
\draw [strokecolor,->] (Jo) ..controls (1415.8bp,860.08bp) and (1390.4bp,823.46bp)  .. (Laurence);
\definecolor{strokecolor}{rgb}{0.15,0.72,0.33};
\draw [strokecolor,->] (Jo) ..controls (1435.2bp,848.55bp) and (1408.3bp,810.2bp)  .. (Laurence);
\definecolor{strokecolor}{rgb}{0.31,0.91,0.92};
\draw [strokecolor,->] (Jo) ..controls (1590.7bp,826.11bp) and (1624.3bp,777.02bp)  .. (Meg);
\definecolor{strokecolor}{rgb}{0.14,0.02,0.05};
\draw [strokecolor,->] (Jo) ..controls (1601.7bp,830.57bp) and (1633.1bp,783.65bp)  .. (Meg);
\definecolor{strokecolor}{rgb}{0.72,0.47,0.95};
\draw [strokecolor,->] (Jo) ..controls (1623.0bp,840.31bp) and (1651.4bp,796.73bp)  .. (Meg);
\definecolor{strokecolor}{rgb}{0.92,0.95,0.18};
\draw [strokecolor,->] (Jo) ..controls (1653.4bp,857.26bp) and (1679.4bp,816.32bp)  .. (Meg);
\definecolor{strokecolor}{rgb}{0.31,0.49,0.64};
\draw [strokecolor,->] (Jo) ..controls (1682.2bp,875.88bp) and (1708.9bp,834.4bp)  .. (Meg);
\definecolor{strokecolor}{rgb}{0.93,0.38,0.46};
\draw [strokecolor,->] (Jo) ..controls (1691.4bp,882.32bp) and (1719.0bp,840.12bp)  .. (Meg);
\definecolor{strokecolor}{rgb}{0.15,0.72,0.33};
\draw [strokecolor,->] (Jo) ..controls (1709.2bp,895.47bp) and (1739.6bp,850.67bp)  .. (Meg);
\definecolor{strokecolor}{rgb}{0.83,0.19,0.37};
\draw [strokecolor,->] (Jo) ..controls (1726.5bp,908.41bp) and (1760.9bp,860.13bp)  .. (Meg);
\definecolor{strokecolor}{rgb}{0.82,0.67,0.29};
\draw [strokecolor,->] (Laurence) ..controls (1261.5bp,960.48bp) and (1290.6bp,1121.2bp)  .. (1319.2bp,1264.0bp) .. controls (1325.4bp,1294.9bp) and (1332.2bp,1327.7bp)  .. (Amy);
\definecolor{strokecolor}{rgb}{0.59,0.25,0.57};
\draw [strokecolor,->] (Laurence) ..controls (1313.8bp,951.1bp) and (1343.8bp,1117.0bp)  .. (1373.2bp,1264.0bp) .. controls (1378.9bp,1292.7bp) and (1385.2bp,1323.0bp)  .. (Amy);
\definecolor{strokecolor}{rgb}{0.47,0.37,0.73};
\draw [strokecolor,->] (Laurence) ..controls (1075.1bp,377.34bp)  .. (Beth);
\definecolor{strokecolor}{rgb}{0.34,0.62,0.08};
\draw [strokecolor,->] (Laurence) ..controls (1338.2bp,846.28bp) and (1364.6bp,883.25bp)  .. (Jo);
\definecolor{strokecolor}{rgb}{0.62,0.45,0.53};
\draw [strokecolor,->] (Laurence) ..controls (1347.7bp,840.38bp) and (1373.4bp,876.81bp)  .. (Jo);
\definecolor{strokecolor}{rgb}{0.24,0.01,0.43};
\draw [strokecolor,->] (Laurence) ..controls (1356.9bp,834.01bp) and (1382.1bp,870.05bp)  .. (Jo);
\definecolor{strokecolor}{rgb}{0.59,0.01,0.23};
\draw [strokecolor,->] (Laurence) ..controls (1366.2bp,827.82bp) and (1391.1bp,863.58bp)  .. (Jo);
\definecolor{strokecolor}{rgb}{0.59,0.25,0.57};
\draw [strokecolor,->] (Laurence) ..controls (1393.2bp,808.12bp) and (1419.4bp,845.44bp)  .. (Jo);
\definecolor{strokecolor}{rgb}{0.35,0.88,0.22};
\draw [strokecolor,->] (Meg) ..controls (1648.2bp,781.42bp) and (1618.1bp,826.44bp)  .. (Jo);
\definecolor{strokecolor}{rgb}{0.72,0.47,0.95};
\draw [strokecolor,->] (Meg) ..controls (1666.4bp,794.61bp) and (1638.9bp,836.96bp)  .. (Jo);
\definecolor{strokecolor}{rgb}{0.24,0.01,0.43};
\draw [strokecolor,->] (Meg) ..controls (1675.8bp,800.94bp) and (1649.2bp,842.36bp)  .. (Jo);
\definecolor{strokecolor}{rgb}{0.86,0.41,0.15};
\draw [strokecolor,->] (Meg) ..controls (1694.9bp,813.51bp) and (1668.8bp,854.4bp)  .. (Jo);
\definecolor{strokecolor}{rgb}{0.79,0.82,0.31};
\draw [strokecolor,->] (Meg) ..controls (1704.7bp,819.56bp) and (1678.6bp,860.52bp)  .. (Jo);
\definecolor{strokecolor}{rgb}{0.64,0.99,0.57};
\draw [strokecolor,->] (Meg) ..controls (1735.0bp,836.58bp) and (1706.4bp,880.09bp)  .. (Jo);
\definecolor{strokecolor}{rgb}{0.15,0.72,0.33};
\draw [strokecolor,->] (Meg) ..controls (1755.9bp,846.83bp) and (1724.1bp,893.46bp)  .. (Jo);
\definecolor{strokecolor}{rgb}{0.73,0.79,0.69};
\draw [strokecolor,->] (Ron) ..controls (608.68bp,930.55bp) and (606.52bp,1136.7bp)  .. (H Potter);
\definecolor{strokecolor}{rgb}{0.41,0.3,0.3};
\draw [strokecolor,->] (Ron) ..controls (636.57bp,934.37bp) and (635.25bp,1134.2bp)  .. (H Potter);
\definecolor{strokecolor}{rgb}{0.07,0.98,0.57};
\draw [strokecolor,->] (Ron) ..controls (664.36bp,936.37bp) and (663.81bp,1133.0bp)  .. (H Potter);
\definecolor{strokecolor}{rgb}{0.04,0.28,0.16};
\draw [strokecolor,->] (Ron) ..controls (678.24bp,937.0bp) and (678.05bp,1132.9bp)  .. (H Potter);
\definecolor{strokecolor}{rgb}{0.41,0.55,0.47};
\draw [strokecolor,->] (Ron) ..controls (692.11bp,937.0bp) and (692.29bp,1132.9bp)  .. (H Potter);
\definecolor{strokecolor}{rgb}{0.3,0.65,0.69};
\draw [strokecolor,->] (Ron) ..controls (747.71bp,932.77bp) and (749.43bp,1135.4bp)  .. (H Potter);
\definecolor{strokecolor}{rgb}{0.2,0.88,0.11};
\draw [strokecolor,->] (Ron) ..controls (761.67bp,930.55bp) and (763.83bp,1136.7bp)  .. (H Potter);
\definecolor{strokecolor}{rgb}{0.41,0.3,0.3};
\draw [strokecolor,->] (Ron) ..controls (604.77bp,383.08bp) .. (Hermione);
\definecolor{strokecolor}{rgb}{0.18,0.68,0.02};
\draw [strokecolor,->] (Snape) ..controls (976.33bp,1203.2bp) and (954.32bp,1235.2bp)  .. (933.17bp,1264.0bp) .. controls (910.59bp,1294.8bp) and (886.03bp,1326.9bp)  .. (H Potter);
\tiny
\matrix [draw,below] at (current bounding box.south) {
  \node [saidnode,label=right:Say] {}; && \node [smilenode,label=right:Smile] {}; && \node [looknode,label=right:Look] {};  &&  \node [emptynode,label=right:Others] {};\\
};
\end{tikzpicture}
\caption{Narrative KGs of HP (left) and LW (right). Nodes corresponds to de-aliased characters and arcs to clustered relations.\label{fig:kg}}
\end{figure}
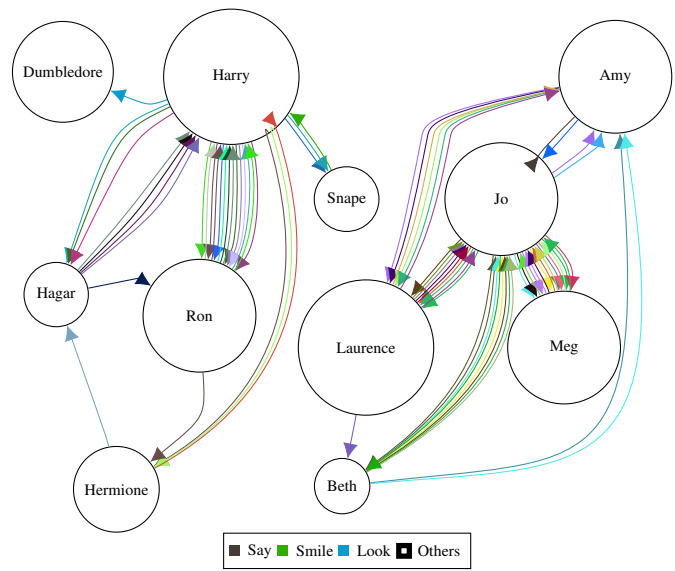

For a first empirical validation of our pipeline we process, in a single run, two novels, namely \textit{Harry Potter and the Philosopher's Stone} (HP) by J. K. Rowling and \textit{Little Women} (LW) by Louisa May Alcott. 1307 suitable sentences out of 32365 are identified and grouped in 200 clusters (i.e. different relations types). As the characters of the two books are distinct, the system generates a KG with two disconnected components (see Figure \ref{fig:kg}). Yet, the relation clustering is able to detect similarities between sentences in the two books. E.g., sentences in Table \ref{tab:sentences} are related to smiling actions. For that cluster the extractive summarization returns the first sentence as a summary and, finally, the triplet generation mechanism return \emph{smile} as representative label.

Concerning sentence clustering we considered both DBSCAN and k-means algorithms both paired with Euclidean and cosine distance. In the considered setup we did not found significant differences with the two metrics. Regarding the algorithms, an observed issue with DBSCAN was a sudden transition from a huge number of single-sentence clusters to very large clusters. Both these extreme scenarios prevent a meaningful identification of relations. Yet, it was not possible to automatically decide the number of clusters with k-means, as the silhouette analysis returned monotone results.

\section{Conclusions}\label{sec:conc}
An unsupervised approach to KG extraction from narrative texts has been proposed. The procedure exploits transformer models to detect similar relations in the triplets, then generates summaries and representative labels for these clusters of similar relations. This represent a pre-processing step for a semi-supervised approach where the representative labels are validated by human annotators and used as a relational classifier. Validated clusters can define relational classifiers, while the automatically generated labels are used for the others. As a future work we want to apply our pipeline to a corpus of literary texts and validate the clusters. This being a starting point for the creation of a knowledge base for literary texts.


\bibliographystyle{named}
\bibliography{ijcai20}
\end{document}